\documentclass[sigconf]{acmart}

\newcommand{\code}[1]{{\color{magenta}#1}}

\usepackage{multirow}
\usepackage[capitalize]{cleveref}
\crefname{section}{Sec.}{Secs.}
\Crefname{section}{Section}{Sections}
\Crefname{table}{Table}{Tables}
\crefname{table}{Tab.}{Tabs.}
\AtBeginDocument{%
  \providecommand\BibTeX{{%
    \normalfont B\kern-0.5em{\scshape i\kern-0.25em b}\kern-0.8em\TeX}}}

\copyrightyear{2022}
\acmYear{2022}
\setcopyright{acmcopyright}\acmConference[MM '22]{Proceedings of the 30th ACM
International Conference on Multimedia}{October 10--14, 2022}{Lisboa, Portugal}
\acmBooktitle{Proceedings of the 30th ACM International Conference on Multimedia
(MM '22), October 10--14, 2022, Lisboa, Portugal}
\acmPrice{15.00}
\acmDOI{10.1145/3503161.3548028}
\acmISBN{978-1-4503-9203-7/22/10}

\begin{document}

\title{Learning Granularity-Unified Representations for Text-to-Image Person Re-identification}

\author{Zhiyin Shao}
\affiliation{%
  \institution{South China University of Technology}
  \city{Guangzhou}
  \country{China}
}
\email{eezyshao@mail.scut.edu.cn}

\author{Xinyu Zhang}
\affiliation{%
  \institution{Baidu VIS}
  \city{Beijing}
  \country{China}
}
\email{zhangxinyu14@baidu.com}

\author{Meng Fang}
\affiliation{%
  \institution{University of Liverpool}
  \city{Liverpool}
  \country{United Kingdom}
}
\email{Meng.Fang@liverpool.ac.uk }

\author{Zhifeng Lin}
\affiliation{%
  \institution{South China University of Technology}
  \city{Guangzhou}
  \country{China}
}
\email{eezhifenglin@mail.scut.edu.cn}

\author{Jian Wang}
\affiliation{%
  \institution{Baidu VIS}
  \city{Beijing}
  \country{China}
}
\email{wangjian33@baidu.com}

\author{Changxing Ding}
\authornote{Corresponding author.}
\affiliation{%
  \institution{South China University of Technology}
  \city{Guangzhou}
  \country{China}
}
\email{chxding@scut.edu.cn}

\renewcommand{\shortauthors}{Zhiyin Shao, et al.}

\begin{abstract}
Text-to-image person re-identification (ReID) aims to search for pedestrian images of an interested identity via textual descriptions. It is challenging due to both rich intra-modal variations and significant inter-modal gaps. Existing works usually ignore the difference in feature granularity between the two modalities, \emph{i.e.}, the visual features are usually fine-grained while textual features are coarse, which is mainly responsible for the large inter-modal gaps. In this paper, we propose an end-to-end framework based on transformers to learn granularity-unified representations for both modalities, denoted as LGUR.
LGUR framework contains two modules: a Dictionary-based Granularity Alignment (DGA) module and a Prototype-based Granularity Unification (PGU) module. In DGA, in order to align the granularities of two modalities, we introduce a Multi-modality Shared Dictionary (MSD) to reconstruct both visual and textual features. 
Besides, DGA has two important factors, \emph{i.e.}, the cross-modality guidance and the foreground-centric reconstruction, to facilitate the optimization of MSD.
In PGU, we adopt a set of shared and learnable prototypes as the queries to extract diverse and semantically aligned features for both modalities in the granularity-unified feature space, which further promotes the ReID performance. 
Comprehensive experiments show that our LGUR consistently outperforms state-of-the-arts by large margins on both CUHK-PEDES and ICFG-PEDES datasets. 
Code will be released at \href{}{\code{https://github.com/ZhiyinShao-H/LGUR}}.

\end{abstract}




\maketitle

\begin{figure}[h]
  \centering
    \includegraphics[width=\linewidth]{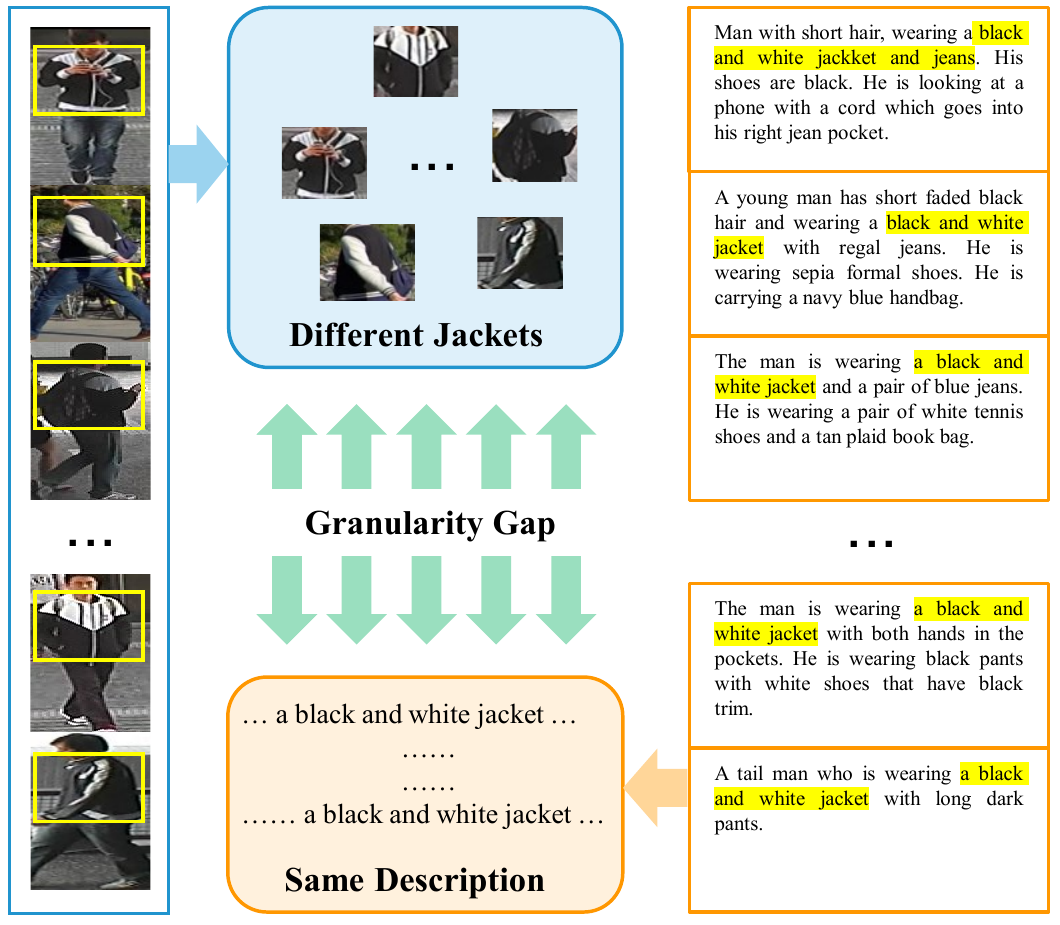}
  \caption{While textual descriptions on the jackets in the all above images are the same, these jackets do in fact differ in terms of their visual details. This example well reflects the granularity gap between the two modalities, \emph{i.e.}, the visual information is fine-grained while the textual features are coarser.}
\label{fig:problems}
\vspace{-0.5cm}
\end{figure}

\section{Introduction}~\label{sec:section1}

Text-to-image person re-identification (ReID) is a cross-modal retrieval task that searches for images of the target identity based on natural language descriptions~\cite{li2017person}. Compared with images, natural language descriptions are more flexible and easier to obtain under certain circumstances. The text-to-image ReID task thus attracts much attention.
However, text-to-image ReID is also significantly more challenging than the image-based ReID~\cite{sun2018beyond, yao2019deep, guo2019beyond, ding2020multi, wang2021batch, wang2022uncertainty, wang2022quality,scheirer2016report} due to the dramatic modality gap between vision and language.

One main aspect of the modality gap relates to the feature granularity. Generally, the visual feature contains \emph{fine-grained} information, while the textual feature describes \emph{coarse} attributes. 
This results in the same textual description being applicable to similar yet different image patches.
For example, in \Cref{fig:problems}, there are several jackets that share the same description ``black and white''; however, these jackets differ in their visual details. This difference in feature granularity enlarges the modality gap and makes the text-to-image ReID more challenging.

The modality gap caused by the feature granularity tends to be ignored by existing works. In fact, the ``granularity gap'' mentioned in the existing text-to-image ReID literature typically refers to the situation in which one word may correspond to image patches of dramatically different sizes~\cite{li2017person,li2017identity,chen2018improving,jing2020pose}. 
These existing approaches do not explicitly solve the modality gap in which an image patch contains more fine-grained information than its corresponding words.
Meanwhile, the common solution in these works is to apply cross-modal attention operations that build the correspondence between image patches and words, facilitating adaptation to the changing size of image patches.

In this paper, we focus on the true granularity gap brought by the fine-grained images and the coarse textual descriptions. We propose a novel Learning Granularity-Unified Representations (LGUR) framework, which maps both visual and textual features into a granularity-unified feature space. The LGUR framework contains a Dictionary-based Granularity Alignment (DGA) module and a Prototype-based Granularity Unification (PGU) module.

In the DGA module, we propose to reconstruct both textual and visual features via a Multi-modality Shared Dictionary (MSD) based on a transformer layer. In the dictionary, we store a set of granularity-unified atoms. The reconstruction operation aims to reduce the feature granularity gap between the two modalities based on these atoms.
Intuitively, the information bottleneck lies in coarse textual features; therefore, the granularity of atoms in MSD should be as close as possible to the textual granularity. 

However, without explicit guidance, it is hard to drive MSD to closely approximate the granularity of textual features.
To address this problem, we introduce the following two strategies in DGA. One is that we guide the learning of MSD parameters using textual features.
More specifically, we reconstruct the visual feature again, using its matched textual feature as value in the same transformer layer as above. 
By reducing the gap between the reconstructed visual features in these two ways, the MSD is forced to be optimized according to the granularity of the textual features. 
Note that we adopt this guidance during training only; therefore, it introduces no additional computational cost in the inference stage.
Another one is that we enable MSD to focus on the foreground pedestrian body. It is because the background is typically ignored by linguistic descriptions and is therefore less relevant to the text-to-image ReID task.
Based on the foreground reconstruction, the optimization difficulty of MSD is significantly reduced.

In the PGU module, we further project the textual and visual features into a unified format by a set of shared and learnable prototypes of one transformer layer. These prototypes extract discriminative and diverse features from the two modalities independently via the cross-attention architecture. Through matching the paired textual and visual features produced by the same prototype, the granularity gap between the two modalities can be further reduced. Meanwhile, thanks to the use of shared prototypes as the query, the computational cost of LGUR is substantially diminished. In comparison, for methods adopting cross-modal attention operations, the visual and textual features are adopted as queries and values in turn; therefore, every image and text must be paired to get the retrieval features, resulting in a heavy computational cost.

We conduct extensive experiments on two existing large-scale benchmark datasets, \emph{i.e.}, CUHK-PEDES \cite{li2017person} and ICFG-PEDES \cite{ding2021semantically}. The results show that our simple LGUR framework consistently and significantly outperforms existing approaches. Compared with many existing methods~\cite{lee2018stacked,niu2020improving,gao2021contextual}, LGUR is also more efficient, since it does not require cross-modal attention operations between each image-text pair in the testing stage. More impressively, we find that LGUR performs well in domain generalization tasks due to the feature unification on the granularity level. 
The main contributions of the proposed method can be summarized as follows:
\begin{itemize}
\item We identify the difference in the feature granularity between the visual and textual modalities that results in the modality gap, which is an important element that is rarely considered in the text-to-image ReID literature.
\item We propose a novel Learning Granularity-Unified Representations (LGUR) framework that efficiently extracts granularity-unified features from both modalities.
\item Extensive experiments on two text-to-image ReID datasets, \emph{i.e.}, CUHK-PEDES and ICFG-PEDES show that LGUR consistently outperforms the state-of-the-arts by large margins.
\end{itemize}

\begin{figure*}[t!] %
  \centering
  \includegraphics[width=1.0\linewidth]{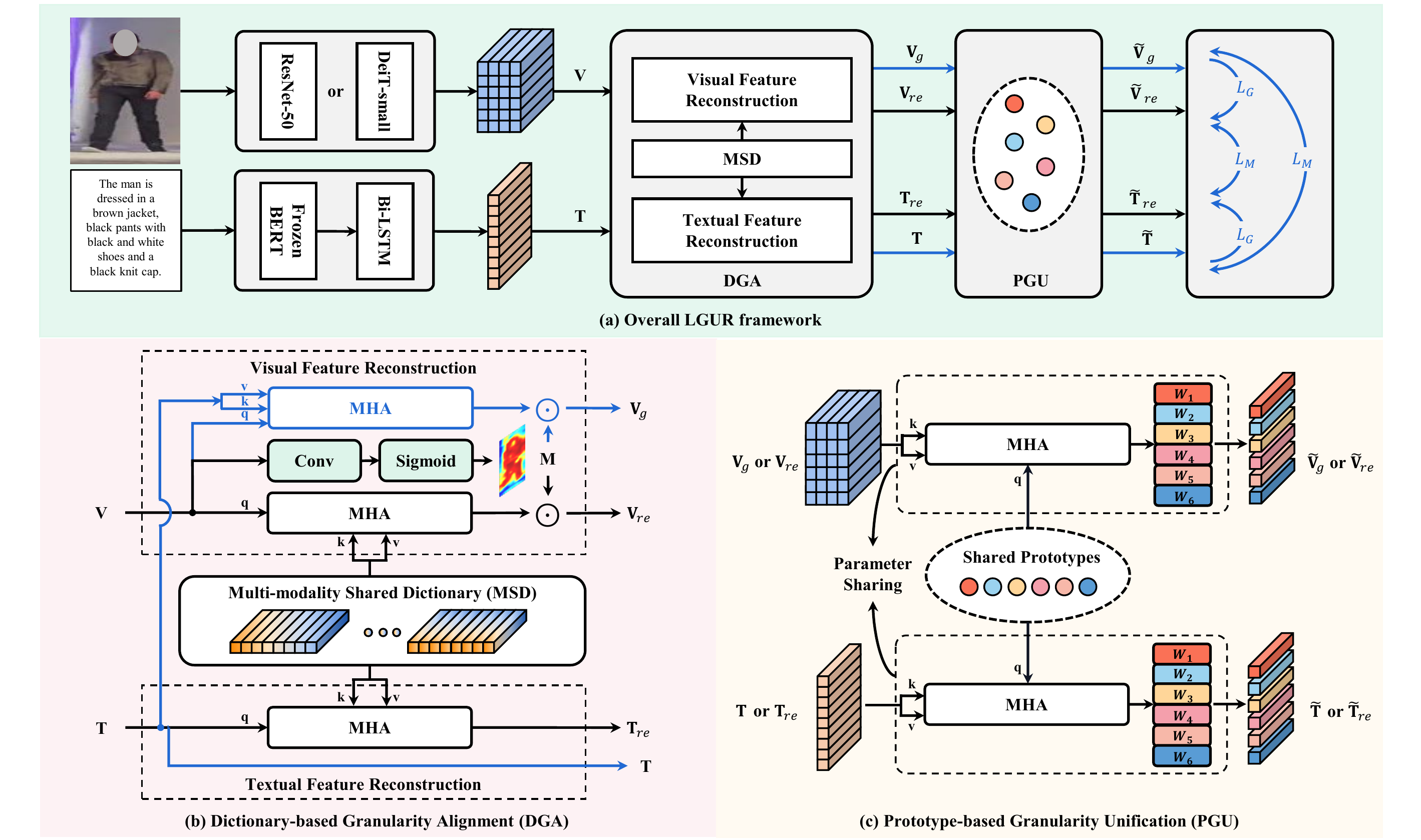}
  \caption{
  Overview of the LGUR framework (shown in a), which includes a Dictionary-based Granularity Alignment (DGA) module (shown in b) and a Prototype-based Granularity Unification (PGU) module (shown in c) to achieve feature extraction that is both efficient and granularity-unified. DGA reconstructs both visual and textual features via a Multi-modality Shared Dictionary (MSD). Moreover, we propose a cross-modal guidance strategy to optimize the MSD parameters according to the granularity of the textual features. In addition, a foreground mask is utilized to enable MSD to focus on the reconstruction of the pedestrian body. PGU projects both textual and visual features into a unified format via a set of shared and learnable prototypes. LGUR does not need to implement any cross-modal attention operations in the testing stage and is therefore computationally efficient. The blue arrows represent operations that are discarded during the testing stage. 
  MHA represents the multi-head attention module. Best viewed in color.
 }
\label{fig:structure}
\end{figure*}

\section{Related Work}
\subsection{Vision-Language Models}\label{sec:setion2.1}

Transformers have demonstrated their superiority on many vision and natural language processing tasks. Multiple works have also been developed that apply the transformer to the vision-language pre-training (VLPT) task \cite{chen2020uniter,li2020unicoder,li2019visualbert,lu2019vilbert,su2019vl,sun2019videobert,tan2019lxmert,zhou2020unified}.
Depending on their model structure, existing VLPT methods can be categorized as either two-stream or single-stream models.
Both types of methods extract vision-language joint features. 
The two-stream models \cite{su2019vl,sun2019videobert,tan2019lxmert,huang2021seeing} extract features from the image and text modalities separately, then fuse them by means of the transformer structure.
For their part, the single-stream models \cite{chen2020uniter,li2020unicoder,li2019visualbert,lu2019vilbert,zhou2020unified} adopt the BERT \cite{devlin2018bert} model and process the image feature and the language feature together as a joint distribution. However, the above approaches require the text and image pair to be fed into the network. Specifically, in the testing stage of 
text-to-image retrieval tasks, every textual query needs to be paired with each image in the gallery, which introduces high computational complexities.

\subsection{Text-to-Image Person ReID}\label{sec:setion2.2}

Due to its fine-grained nature, text-to-image person ReID is more challenging than general cross-modal retrieval tasks. Depending on the alignment strategy utilized, existing works can be divided into cross-modal attention-based methods and cross-modal attention-free methods.
The latter type of methods design various model structures \cite{zheng2020dual,wang2020vitaa} or objective functions \cite{li2017identity,faghri2017vse++,sarafianos2019adversarial,zhang2018deep,liu2019deep,ge2019visual} to align the features from both modalities in a shared feature space.

In comparison, cross-modal attention-based methods focus on establishing region-word \cite{li2017person,li2017identity,chen2018improving} or region-phrase \cite{jing2020pose,niu2020improving} correspondences. 
These methods have their own advantages as well as disadvantages. Cross-modal attention-free methods are usually more efficient. Specifically, given $N$ images and $M$ sentences, their complexities are $O(M+N)$. By contrast, the complexity of the cross-modal attention-based methods increases to $O(MN)$ \cite{wang2018learning}. However, these methods usually achieve significantly better retrieval performance as they better reduce the modality gaps. 

The granularity gap mentioned in existing cross-modal attention-based methods \cite{niu2020improving,wang2021text,jing2020pose} usually refers to the situation in which each word may correspond to image patches of dramatically different sizes. However, these methods rarely pay attention to the modality gap in feature granularity, \emph{i.e.}, similar but different image regions may share the same textual description. In this paper, we handle this new problem by means of a novel LGUR framework. As LGUR avoids cross-modal attention operations between image and text through the use of a modality-shared dictionary, it has great advantages in computational efficiency. 

\section{Methodology}
The overview of our LGUR framework is 
illustrated in \Cref{fig:structure}. 
LGUR comprises three modules: the Feature Extraction Backbones (see Section~\ref{sec:setion3.1}), the Dictionary-based Granularity Alignment Module (see Section~\ref{sec:setion3.2}) and the Prototype-based Granularity Unification Module (see Section~\ref{sec:setion3.3}).
The latter two modules are used to enhance the granularity unification of the image and text modalities.
The optimization of the overall framework is described in \Cref{sec:setion3.4}.

\subsection{Feature Extraction Backbones}\label{sec:setion3.1}
\noindent\textbf{Visual modality.}
Let ${\bf{V}}\in\mathbb{R}^{HW \times d}$ represents the visual feature produced by the visual backbone, while
$d$ denotes the feature dimension.
We consider two backbones, namely DeiT-Small~\cite{touvron2021training} and ResNet50~\cite{he2016deep}.
For DeiT-Small, we split the image into ${H\times W}$ patches, so that $HW$ denotes the number of patch tokens.
As for ResNet50, $H$ and $W$ are the height and width respectively of its output feature maps. These two backbones have a similar number of parameters.

\noindent\textbf{Textual modality.}
Let ${\bf{T}}\in\mathbb{R}^{L \times d}$ represents the output of the textual backbone.
We adopt a light-weight bidirectional long short-term memory network (Bi-LSTM) \cite{hochreiter1997long} to extract textual features.
The input embeddings of Bi-LSTM are obtained from a pretrained BERT model \cite{devlin2018bert}.
$L$ stands for the number of words in the description. The backbones are introduced in more details in Section~\ref{sec:section4.2}.

\subsection{Dictionary-based Granularity Alignment}\label{sec:setion3.2}
As analyzed in Section~\ref{sec:section1}, the textual granularity is coarse while the visual granularity is fine-grained. This large granularity gap creates inconsistency between the two types of feature representations.
We conjecture that the visual and textual features will match with each other more tightly if the former are made to be more abstract.
To this end, we introduce a Dictionary-based Granularity Alignment (DGA) module to \emph{reconstruct} the representations of both modalities, such that these features can be made consistent in a granularity-unified feature space.
Our DGA utilizes a Multi-modality Shared Dictionary (MSD) to conduct the textual and visual feature reconstruction.

\noindent\textbf{Multi-modality shared dictionary.} 
We build the MSD as ${\bf{D}}\in \mathbb{R}^{s \times d}$, which is randomly initialized.
Here, $s$ indicates the number of atoms in ${\bf{D}}$. 
Each atom in ${\bf{D}}$ is a $d$-dimensional vector, which has the same dimension as ${\bf{V}}$ and ${\bf{T}}$.
We expect ${\bf{D}}$ to possess similar granularity to the textual features.
In the following, we describe the way in which granularity-aligned visual and textual features can be obtained via reconstruction using ${\bf{D}}$.

\noindent\textbf{Textual feature reconstruction.}
We first apply ${\bf{D}}$ to reconstruct textual features.
The textual features before and after reconstruction are expected to be similar with each other. To this end, we minimise the similarity of these two textual features using a ranking loss, 
which will be described in Section~\ref{sec:setion3.4}. This strategy drives the atoms in ${\bf{D}}$ to possess similar granularity to that of the text.

Formally, we utilize a transformer’s cross-attention operation as the reconstruction process \cite{vaswani2017attention}, in which ${\bf{T}}$ is utilized as the query while ${\bf{D}}$ acts as the key and value.
The reconstructed textual feature ${\bf{T}}_{re}$ can be expressed as follows:
\begin{equation}
{\bf{T}}_{re} = MHA_1({\bf{T}}, {\bf{D}}, {\bf{D}}), %
\label{eq:textual_reconstruction}
\end{equation}
where $MHA_1(\cdot)$ denotes a transformer block, which consists of a multi-head attention and a feed-forward network \cite{vaswani2017attention}.
More formally,
$MHA({\bf{Q}}, {\bf{K}}, {\bf{V}}) = FFN(MultiHead({\bf{Q}}, {\bf{K}}, {\bf{V}}))$,
where ${\bf{Q}}$, ${\bf{K}}$, and ${\bf{V}}$ are abbreviations for query, key, and value, respectively.

\noindent\textbf{Visual feature reconstruction.}
We also reconstruct the visual features via $MHA_1(\cdot)$ with ${\bf{V}}$ as the query and ${\bf{D}}$ as the key and value. The reconstructed visual feature is denoted as ${\bf{V}}_{re}\in\mathbb{R}^{HW \times d}$.

Different from the textual feature reconstruction, we propose the following two strategies in the visual reconstruction to further reduce the modality gap between ${\bf{V}}_{re}$ and ${\bf{T}}_{re}$.
First, we enable MSD to focus on the reconstruction of the image foreground, \emph{i.e.}, the pedestrian body.
This is based on the consideration that pedestrian images are usually characterized by rich occlusion and background clutters, while textual descriptions are identity-centric and tend to ignore the background noises. 
Therefore, we generate a foreground mask ${\bf{M}}\in\mathbb{R}^{HW \times 1}$ via the spatial attention mechanism \cite{song2018mask}. 
Specifically, we attach a 1$\times$1 convolutional layer with a sigmoid function to $\bf{V}$ to obtain $\bf{M}$. 
As illustrated in Figure~\ref{fig:structure}, the reconstructed visual features from MSD based on the foreground restriction is shown as follows:
\begin{equation}
{\bf{V}}_{re} = MHA_1({\bf{V}}, {\bf{D}}, {\bf{D}}){\odot}\bf{M},
\label{eq:visual_reconstruction}
\end{equation}
where $\odot$ represents the Hadamard product between $\bf{M}$ and each column of $MHA_1({\bf{V}}, {\bf{D}}, {\bf{D}})$. With the help of $\bf{M}$, the reconstructed background clutters are suppressed and the optimization difficulty of MSD is alleviated.

Second, we guide the granularity of ${\bf{V}}_{re}$ to be abstract under the help of the textual features. More specifically, we reconstruct $\bf{V}$ again using $\bf{T}$ from the paired language description with the input image. $\bf{T}$ acts as both the key and value in the transformer block:
\begin{equation}
{\bf{V}}_{g} = MHA_1({\bf{V}},{\bf{T}}, {\bf{T}}){\odot}\bf{M},
\label{eq:visual_foreground_guidance}
\end{equation}
where ${\bf{V}}_{g}\in\mathbb{R}^{HW \times d}$.  

Compared with ${\bf{V}}_{re}$, the granularity of ${\bf{V}}_{g}$ is closer to that of textual features. 
Therefore, we impose a ranking loss to penalize the difference between ${\bf{V}}_{re}$ and ${\bf{V}}_{g}$, as detailed in Section~\ref{sec:setion3.4}.

\subsection{Prototype-based Granularity Unification}\label{sec:setion3.3}
In Section~\ref{sec:setion3.2}, we align the granularity between image and text via MSD. Despite this, the model’s ability to accurately match texts and images of a specific identity remains limited. In fact, what $\bf{D}$ has learned is general semantic knowledge. In this subsection, we aim to extract more powerful features for the ReID purpose via a Prototype-based Granularity Unification (PGU) module.
PGU projects both textual and visual features into a unified format, which further aligns the granularity of both modalities.
More specifically, we design a set of prototypes ${\bf{P}} = [{\bf{p}}_1, {\bf{p}}_2, \cdots, {\bf{p}}_K]\in\mathbb{R}^{d \times K}$, which are randomly initialized.
The $K$ prototypes contain diverse semantic information.
To enable these prototypes to capture both textual and visual features, we let each prototype act as the query in the transformer layer, while textual or visual feature acts as both the key and value.
For simplicity, we define $\bf{F}$ as an example to represent a textual or visual feature. 
As shown in Figure~\ref{fig:structure}, the refined feature $\widetilde{{\bf{F}}}\in \mathbb{R}^{K \times d^{'}}$ after PGU is defined as follows:
\begin{equation}
\begin{aligned}
\widetilde{\bf{F}} = PGU({\bf{P}},{\bf{F}}) &= Concat(f_1({\bf{p}}_1, {\bf{F}}),\dots,f_K({\bf{p}}_K, {\bf{F}})), \\
f_i({\bf{p}}_i, {\bf{F}}) &= {{\bf{W}}_{k}}(MHA_2({{\bf{p}}_i},{\bf{F}},{\bf{F}})),
\end{aligned}
\label{eq:PGU}
\end{equation}
where ${\bf{W}}_{k}\in \mathbb{R}^{d^{'} \times d}$ denotes an FC layer for the $k$-th query ${\bf{p}}$.
We apply independent FC layers for the queries in order to produce semantically diverse features.
Meanwhile, each query adopts the same FC layer for both modalities, which further aligns their feature granularity.
$d^{'}$ is the output dimension after FC.
$Concat$ denotes the concatenation operation.
$MHA_2$ is of the same structure as $MHA_1$ in Eq.~\eqref{eq:textual_reconstruction} albeit with independent parameters.
Based on Eq.~\eqref{eq:PGU}, we obtain the granularity-unified feature
$\widetilde{{\bf{T}}}_{re}$ of ${\bf{T}}_{re}$ in Eq.~\eqref{eq:textual_reconstruction},
$\widetilde{\bf{V}}_{re}$ of ${\bf{V}}_{re}$ in Eq.~\eqref{eq:visual_reconstruction}, 
$\widetilde{\bf{T}}$ of ${\bf{T}}$, and 
$\widetilde{\bf{V}}_{g}$ of ${\bf{V}}_{g}$ in Eq.~\eqref{eq:visual_foreground_guidance}. More formally,
\begin{equation}
\begin{aligned}
\widetilde{{\bf{T}}}_{re} &= PGU({\bf{P}},{{\bf{T}}_{re}}),
\widetilde{\bf{V}}_{re} = PGU({\bf{P}},{{\bf{V}}_{re}}), \\
\widetilde{\bf{T}} &= PGU({\bf{P}},{{\bf{T}}}),
\widetilde{\bf{V}}_{g} = PGU({\bf{P}},{{\bf{V}}_{g}}).
\end{aligned}
\label{eq:PGU_all}
\end{equation}
Here, we use a single shared $MHA_2$ in Eq.~\eqref{eq:PGU_all}.
After PGU, the granularities of the visual and textual features are aligned to a unified space, which substantially reduces the granularity gap.

\noindent\textbf{Discussion.}
Previous works \cite{li2017person,li2017identity,chen2018improving,jing2020pose,niu2020improving,gao2021contextual} usually rely on a cross-model attention between two modalities to reduce modality gaps.
However, all image-text pairs need to be fed into this models of this kind, which is very computationally costly.
In comparison, the prototypes in our PGU are unified representations for both modalities.
For a single piece of text or image, we directly obtain the feature $\widetilde{{\bf{T}}}_{re}$ or $\widetilde{\bf{V}}_{re}$ rather than first composing text-to-image pairs; thus, our PGU is more computationally efficient.
Meanwhile, the format of our prototype is similar to the object query in \cite{carion2020end}.
However, in \cite{carion2020end}, the output corresponding to each object query represents a potential object instance, while the output corresponding to each prototype in our PGU denotes a discriminative region of one pedestrian.
The prototypes in PGU thus contain more detailed information.

\subsection{Optimization \& Inference}\label{sec:setion3.4}

\noindent\textbf{Optimization.} 
Inspired by \cite{sun2018beyond,zhang2018deep}, we adopt cross-entropy loss as the identification loss for each prototype.
For a specific feature $\widetilde{\bf{F}}$ in Eq.~\eqref{eq:PGU}, we denote the predicted identity probabilities by the \emph{k}-th prototype as $\hat{\bf{y}}_{k}$. The identification loss can thus be written as:
\begin{equation}
L_{ID}(\widetilde{\bf{F}}) = {\dfrac{1}{K}} \sum_{k=1}^K -{\bf{y}}~{\odot}~log (\hat{{\bf{y}}}_k),
\label{eq:id_loss}
\end{equation}
where ${\bf{y}}$ is the ground-truth label vector. 

Meanwhile, ranking loss is commonly applied to the text-to-image ReID task.
For two features $\widetilde{\bf{F}}_1$ and $\widetilde{\bf{F}}_2$ from one matched image-text pair, the ranking loss is formulated as follows:
\begin{equation}
\begin{aligned}
L_{RK}(\widetilde{\bf{F}}_1 , \widetilde{\bf{F}}_2)
&= max(\alpha - S(\widetilde{\bf{F}}_1 , \widetilde{\bf{F}}_2^{+}) + S(\widetilde{\bf{F}}_1, \widetilde{\bf{F}}_2^{-}),0)\\
&+max(\alpha - S(\widetilde{\bf{F}}_2, \widetilde{\bf{F}}_1^{+}) + S(\widetilde{\bf{F}}_2, \widetilde{\bf{F}}_1^{-}),0),
\end{aligned}
\label{eq:tri_loss}
\end{equation}
where $\widetilde{\bf{F}}_1^{+}$/$\widetilde{\bf{F}}_2^{+}$ and $\widetilde{\bf{F}}_1^{-}$/$\widetilde{\bf{F}}_2^{-}$ are one positive sample and one semi-hard negative sample of $\widetilde{\bf{F}}_1$/ $\widetilde{\bf{F}}_2$ in a mini-batch, respectively.
In addition, $\alpha$ is a margin hyper-parameter, while $S$ denotes the cosine similarity metric.

By applying Eq.~\eqref{eq:id_loss} and Eq.~\eqref{eq:tri_loss} to LGUR, we obtain the following loss:
\begin{equation}
\begin{aligned}
L_{M} &= L_{ID}(\widetilde{\bf{T}}_{re}) + L_{ID}(\widetilde{\bf{V}}_{re}) + L_{ID}(\widetilde{\bf{T}}) + L_{ID}(\widetilde{\bf{V}}_{g}) \\
& + L_{RK}(\widetilde{\bf{T}}_{re}, \widetilde{\bf{V}}_{re}) + L_{RK}(\widetilde{\bf{T}} , \widetilde{\bf{V}}_{g}).
\end{aligned}
\end{equation}

Moreover, to achieve tighter granularity alignment, we impose another loss function based on the guidance features, $\widetilde{\bf{V}}_{g}$ and $\widetilde{\bf{T}}$, as described in Section~\ref{sec:setion3.2}. 
Specifically, we adopt the ranking loss to pull the reconstructed features closer to the guidance features when they refer to the same person, or push them away when they refer to different identities.
The guidance loss can be represented as follows:
\begin{equation}
L_{G} = L_{RK}(\widetilde{\bf{T}}_{re}, \widetilde{\bf{T}}) + L_{RK}(\widetilde{\bf{V}}_{re}, \widetilde{\bf{V}}_{g}).
\end{equation}
The overall loss function can thus be expressed as:
\begin{equation}
L = L_{M} + L_{G}.
\end{equation}

\noindent\textbf{Inference.} We separately extract the textual feature $\widetilde{{\bf{T}}}_{re}$ and the visual feature $\widetilde{{\bf{V}}}_{re}$ for the text-to-image retrieval.
$\widetilde{\bf{T}}$ and $\widetilde{\bf{V}}_{g}$ are abandoned during inference. The cosine similarity is adopted as the metric for retrieval.

\begin{table}[t!]
\centering
\caption{Performance comparisons on CUHK-PEDES.}
\begin{tabular} {c|c|ccc}
    \hline
      \multicolumn{2}{c|}{Methods}  & Rank-1 & Rank-5 & Rank-10 \\
    \hline
      \multirow{19}*{\rotatebox{90}{ResNet-50}}
       &  Dual Path \cite{zheng2020dual}  & 44.40 & 66.26 & 75.07\\
       & CMPM/C \cite{zhang2018deep}  & 49.37 & - & 79.27\\
       & MIA \cite{niu2020improving}  &53.10 & 75.00 & 82.90\\
       & A-GANet \cite{liu2019deep}  & 53.14 & 74.03 & 82.95\\
       & GALM \cite{jing2020pose}  & 54.12 & 75.45 & 82.97\\
       & TIMAM \cite{sarafianos2019adversarial}  & 54.51 & 77.56 & 84.78\\
       & TDE \cite{niu2020textual}  & 55.25 & 77.46 & 84.56\\
       & VTA \cite{ge2019visual}  & 55.32 & 77.00 & 84.26\\
       & SCAN \cite{lee2018stacked} &  55.86 & 75.97 & 83.69\\
       & ViTAA \cite{wang2020vitaa}  & 55.97 & 75.84 & 83.52\\
       & CMAAM \cite{aggarwal2020text}  & 56.68 & 77.18 & 84.86\\
       &  HGAN \cite{zheng2020hierarchical}  & 59.00 & 79.49 & 86.62\\
       &  NAFS \cite{gao2021contextual} & 59.94 & 79.86 & 86.70\\
       &  DSSL \cite{zhu2021dssl} & 59.98 & 80.41 & 87.56\\
       &  MGEL \cite{wang2021text} &  60.27 & 80.01 & 86.74\\
       &  SSAN \cite{ding2021semantically}  & 61.37 & 80.15 & 86.73\\
      &  {Han \textit{et al.}} \cite{han2021text} &{61.65} &{80.98} &{86.78} \\
      &  LapsCore \cite{wu2021lapscore} & 63.40 & - & 87.80\\
      &  \bfseries LGUR  &{\bfseries 64.21} &{\bfseries 81.94} &{\bfseries 87.93} \\
    \hline 
    \multicolumn{2}{c|}{{\bfseries LGUR} ( DeiT-Small)}  &{\bfseries 65.25} &{\bfseries 83.12} &{\bfseries 89.00} \\
    \hline
\end{tabular}
\label{tab:CUHKCompare}
\end{table}
\section{Experiments}
We evaluate the LGUR framework on two datasets, namely CUHK-PEDES and ICFG-PEDES. We further adopt the Rank-1, Rank-5, and Rank-10 accuracies as metrics to evaluate performance on both databases.

\subsection{Datasets and Evaluation Metrics}
\noindent\textbf{CUHK-PEDES.} CUHK-PEDES~\cite{li2017person} contains 40,206 images and 80,412 textual descriptions for 13,003 identities.
The training set comprises 34,050 images and 68,120 textual descriptions of 11,000 pedestrians.
The testing  set includes 3,074 images and 6,156 textual descriptions of the rest 1,000 pedestrians. 
Each image contains at least two textual descriptions, each of which is made up of 23.5 words on average.

\noindent\textbf{ICFG-PEDES.} ICFG-PEDES \cite{ding2021semantically} contains 54,522 pedestrian images of 4,102 different identities,  all of which were collected from the MSMT17 database \cite{wei2018person}. 
The training set includes 34,674 images of 3,102 pedestrians. The testing set comprises 19,848 images of 1,000 pedestrians. Each image is associated with only one textual description; these descriptions contain 37.2 words on average.

\noindent\textbf{Evaluation metrics.} We adopt the popular Rank-$k$ metrics ($k$=1,5,10) as the evaluation metrics. 
Rank-$k$ reveals the probability that, when given a textual description as query, we can find at least one matching person image in the top-$k$ candidate list.

\begin{table}[t!]
\centering
\caption{Performance comparisons on ICFG-PEDES.}
\begin{tabular}{c|c|ccc}
    \hline
      \multicolumn{2}{c|}{Methods}  & Rank-1 & Rank-5 & Rank-10 \\
    \hline
        \multirow{7}*{\rotatebox{90}{ResNet-50}}
       & Dual Path \cite{zheng2020dual}  & 38.99 & 59.44 & 68.41\\
       & CMPM/C \cite{zhang2018deep} & 43.51 & 65.44 & 74.26\\
       & MIA \cite{niu2020improving}  &46.49 & 67.14 & 75.18\\
       & SCAN \cite{lee2018stacked} & 50.05 & 69.65 & 77.21\\
       & ViTAA \cite{wang2020vitaa} & 50.98 & 68.79 & 75.78\\
       & SSAN \cite{ding2021semantically} & 54.23 & 72.63 & 79.53\\
       & \bfseries LGUR &{\bfseries 57.42} &{\bfseries 74.97} &{\bfseries 81.45} \\
       \hline
       \multicolumn{2}{c|}{{\bfseries LGUR} ( DeiT-Small)}  &{\bfseries 59.02} &{\bfseries 75.32} &{\bfseries 81.56} \\
       
    \hline
\end{tabular}
\label{tab:ICFGCompare}
\end{table}

\begin{table}[t!]
\centering
\caption{Performance comparisons on the domain generalization task. ``C'' denotes CUHK-PEDES, while ``I'' represents ICFG-PEDES.}
\begin{tabular}{c|c|ccc}
    \hline
      \multicolumn{2}{c|}{Methods} &Rank-1 & Rank-5 & Rank-10 \\
    \hline
    \hline
      \multirow{6}*{\rotatebox{90}{C $\rightarrow$ I}}
       & Dual Path \cite{zheng2020dual} & 15.41 & 29.80 & 38.19\\
       & MIA \cite{niu2020improving} & 19.35 & 36.78 & 46.42\\
       & SCAN \cite{lee2018stacked} & 21.27 & 39.26 & 48.83\\
       & SSAN \cite{ding2021semantically} & 24.72 & 43.43 & 53.01\\
       & SSAN(w/ BERT) \cite{ding2021semantically} & 29.24 & 49.00 & 58.53\\
      \cline{2-5}
      & \bfseries LGUR &{\bfseries 34.25} &{\bfseries 52.58} &{\bfseries 60.85} \\

    \hline\hline
      \multirow{6}*{\rotatebox{90}{I $\rightarrow$ C}}
        & Dual Path \cite{zheng2020dual} & 7.63 & 17.14 & 23.52\\
       & MIA \cite{niu2020improving} & 10.93 & 23.77 & 32.39\\
       & SCAN \cite{lee2018stacked} & 13.63 & 28.61 & 37.05\\
       & SSAN \cite{ding2021semantically} & 16.68 & 33.84 & 43.00\\
       & SSAN(w/ BERT) \cite{ding2021semantically} & 21.07 & 38.94 & 48.54\\
      \cline{2-5}
      & \bfseries LGUR &{\bfseries 25.44} &{\bfseries 44.48} &{\bfseries 54.39} \\
    \hline
\end{tabular}
\label{tab:crossdomain}
\end{table}

\subsection{Implementation Details}\label{sec:section4.2}

In our experiments, we choose to use DeiT-Small \cite{touvron2021training} with a patch size of 16 and ResNet-50~\cite{he2016deep} as the visual backbones, respectively. We attach one 1$\times$1 convolutional layer to the visual backbone to project its output to $d$-dim.
For the textual modality, the sequence of word embeddings extracted from BERT is then fed to a Bi-LSTM. Note that we ``freeze'' the weights of BERT, similar to \cite{sarafianos2019adversarial}, and only fine-tune Bi-LSTM.
We resize all images to 384 $\times$ 128 pixels and use only random horizontal flipping as the data augmentation. We set the feature dimension $d$ for both the image and text to 384.
$d^{'}$ is set to 512. The dictionary size $s$ is 400 and the margin $\alpha$ is set to 0.3.
The number of shared prototypes $K$ is set to 6.
During training, we adopt the Adam optimizer \cite{kingma2014adam}.
The batch size is 64, and the number of epochs is 60. 
The initial learning rate of DeiT-Small is set to 0.0001, while the others are set to 0.001.

\begin{table}[t!]
\centering
\caption{Performance comparisons in terms of time complexity. ``CAM'' refers to the cross-modal attention mechanism~\cite{lee2018stacked}.}
\begin{tabular}{c|c|c|ccc}
    \hline
      & CAM & \multicolumn{1}{c|}{Methods} & Train & Inference & Rank-1 \\
    \hline
    \hline
      \multirow{7}*{\rotatebox{90}{CUHK-PEDES}} &
     \checkmark & MIA \cite{niu2020improving} & 680ms & 42ms &53.10\\
      & \checkmark & SCAN \cite{lee2018stacked} & 718ms & 46ms &55.86\\
      & \checkmark & NAFS \cite{gao2021contextual} & 1,284ms & 42ms &59.94\\
       \cline{2-6}
      & $\times$ & Dual Path \cite{zheng2020dual} & 321ms & 10ms &44.40\\
      & $\times$ & CMPM/C \cite{zhang2018deep} & 338ms & 27ms &49.37\\
      & $\times$ & SSAN \cite{ding2021semantically} & 901ms & 76ms &61.37\\
      & $\times$ & LGUR (Ours) & 886ms & 26ms &64.21 \\

    \hline\hline
      \multirow{7}*{\rotatebox{90}{ICFG-PEDES}}
     & \checkmark  & MIA \cite{niu2020improving} & 711ms & 113ms &46.49\\
     & \checkmark  & SCAN \cite{lee2018stacked} & 738ms & 114ms &50.05\\
     & \checkmark  & NAFS \cite{gao2021contextual} & 1,304ms & 116ms &-\\
       \cline{2-6}
     & $\times$  & Dual Path \cite{zheng2020dual} & 342ms & 11ms &38.99\\
     & $\times$  & CMPM/C \cite{zhang2018deep} & 356ms & 31ms &43.51\\
     & $\times$  & SSAN \cite{ding2021semantically} & 973ms & 77ms &54.23\\
     & $\times$ & LGUR (Ours) & 910ms & 31ms &57.42\\
    \hline
\end{tabular}
\label{tab:costtime}
\end{table}

\subsection{Comparisons with State-of-the-Art Methods}
To facilitate fair comparison, we evaluate the performance of LGUR with DeiT-Small and ResNet-50 as the visual backbone, respectively.

\noindent\textbf{Comparisons on CUHK-PEDES.}
In Table~\ref{tab:CUHKCompare}, our LGUR outperforms all state-of-the-art methods, achieving Rank-1 accuracy of 64.21\% and 65.25\% based on ResNet-50 and DeiT-Small, respectively.
In particular, our LGUR outperforms NAFS \cite{gao2021contextual} (which also adopts BERT for textual feature extraction) by as much as 4.27\% in terms of Rank-1 accuracy.
Moreover, NAFS requires cross-modality attention operations, which are computationally expensive.
By contrast, LGUR extracts textual and visual features independently and thereby substantially reduces the computational cost (as discussed in Section~\ref{sec:setion3.3}).
LGUR also achieves higher performance than one most recent method, named LapsCore~\cite{wu2021lapscore}.
It is worth noting that LapsCore is based on the NAFS~\cite{gao2021contextual} model and therefore also has a much higher computational cost than LGUR.
Furthermore, LapsCore focuses on designing auxilary tasks for regularization and does not consider the granularity gap between the two modalities; therefore, the contributions of LGUR and LapsCore complement each other.

\noindent\textbf{Comparisons on ICFG-PEDES.}
Comparison results are summarized in Table~\ref{tab:ICFGCompare}.
Since ICFG-PEDES is a new database, we directly cite the performance of existing approaches evaluated in \cite{ding2021semantically}.
LGUR consistently achieves the best performance.
Specifically, it achieves 57.42\% and 59.02\% Rank-1 accuracies with the ResNet-50 and DeiT-Small backbones, respectively.
SSAN \cite{ding2021semantically} achieves superior performance since it extracts fine-grained part-level textual and visual features.
However, this method still ignores the granularity gap between the two modalities.
By bridging the granularity gap, LGUR outperforms SSAN by 3.19\% in terms of Rank-1 accuracy.

\noindent\textbf{Comparisons on the domain generalization (DG) task.}
Our LGUR effectively narrows the granularity gap between the textual and visual features.
Due to the feature unification on a coarse granularity level,
it could be naturally assumed that the model is able to generalize well to the other domains.
To this end, we conduct experiments on DG tasks.
Here, we directly deploy the model pretrained on the source domain to the target dataset.
As shown in Table~\ref{tab:crossdomain}, our LGUR outperforms all other comparison methods.
In particular, LGUR achieves Rank-1 improvements of 9.53\% and 8.76\% on the C$\rightarrow$I and I$\rightarrow$C settings respectively when compared to SSAN~\cite{ding2021semantically}.
To exclude the differences arising from BERT~\cite{devlin2018bert}, we also equip SSAN with the same textual feature extraction backbone as LGUR.
The performance of SSAN increases, yet is still lower than that of LGUR by 5.01\% and 4.37\% in terms of Rank-1 accuracy.
This experiment demonstrates that the granularity-unified representations of text and image have good generalization ability.

\begin{table*}[t!]
\centering
\caption{Ablation study on key modules of LGUR. 
DGA has three important components: the multi-modality shared dictionary ($\bf{D}$), the  foreground mask ($\bf{M}$), the guidance ($G$) for reconstruction (${\bf{T}}$ and ${\bf{V}}_g$).
}
\begin{tabular}{c|cc|c|ccc|ccc|ccc}
\hline
   \multirow{3}*{No.} & \multicolumn{2}{c|}{\multirow{3}*{Methods}} & \multicolumn{4}{c|}{Components} & \multicolumn{3}{c|}{CUHK-PEDES} & \multicolumn{3}{c}{ICFG-PEDES} \\
  \cline{4-13}
  &  & & \multirow{2}*{PGU}  &\multicolumn{3}{c|}{DGA}
  & \multirow{2}*{Rank-1}  & \multirow{2}*{Rank-5} & \multicolumn{1}{c|}{\multirow{2}*{Rank-10}} & \multirow{2}*{Rank-1} & \multirow{2}*{Rank-5} & \multirow{2}*{Rank-10} \\
  \cline{5-7}
  & & & & $\bf{D}$ & $\bf{M}$  & $G$  &\multicolumn{3}{c|}{}\\
  \hline
  0 & \multicolumn{2}{c|}{baseline}  &-    &-         &-              & -             & 58.67 & 79.08 & 85.82 & 52.09 & 70.91 & 78.07\\
  1 & \multicolumn{2}{c|}{+ PGU} &\checkmark    &-         &-              & -             & 63.26 & 81.17 & 87.69 & 56.34 & 73.58 & 80.23\\
  2 & \multicolumn{2}{c|}{+ DGA}             &-      &\checkmark     &\checkmark     &\checkmark    & 61.86 & 80.43 & 87.20  &55.83         &73.65              & 80.48\\
  \hline
  3 & \multicolumn{2}{c|}{+ PGU + DGA (only $\bf{D}$) }      &\checkmark       &\checkmark     &-              & -             & 64.28 & 81.95 & 88.11 & 57.52 & 74.84 & 81.14\\
  4 & \multicolumn{2}{c|}{+ PGU + DGA ($\bf{D}$ + $\bf{M}$)}        &\checkmark   &\checkmark     &\checkmark  &  -   & 64.64 & 82.91 & 88.52 & 57.82 & 74.99 & 81.17\\
  5 & \multicolumn{2}{c|}{+ PGU + DGA ($\bf{D}$ + G)}        &\checkmark  &\checkmark     & -     & \checkmark             & 64.80 & 82.29 & 88.32 & 58.17 & 75.83 & 81.51\\
    \hline
  6 &  \multicolumn{2}{c|}{LGUR}               &\checkmark      &\checkmark     &\checkmark     &\checkmark    &{\bfseries 65.25} &{\bfseries 83.12} &{\bfseries 89.00}  &\bfseries 59.02 & \bfseries 75.32 & \bfseries 81.56\\
  \hline
\end{tabular}
\label{tab:ablation}
\end{table*}

\begin{table}
\centering
\caption{Comparisons with variants of MSD, including without reconstruction (w/o reconstruction), reconstruction with a self-attention layer (w/ SA)~\cite{vaswani2017attention}, reconstruction with modality unshared dictionary (w/ unshared ${\bf{D}}$).
}
\begin{tabular}{c|cc|cc}
    \hline
      \multirow{2}*{Type of reconstruction} & \multicolumn{2}{c|}{CUHK-PEDES} & \multicolumn{2}{c}{ICFG-PEDES}\\
    \cline{2-5}
       & \multicolumn{1}{c}{Rank-1} & \multicolumn{1}{c|}{Rank-5}& \multicolumn{1}{c}{Rank-1} & \multicolumn{1}{c}{Rank-5}\\
    \hline
        w/o reconstruction & 63.26  & 81.17 & 56.34  & 73.58\\
        w/ SA & 63.39 & 82.28 & 56.34 & 74.09\\
        w/ unshared ${\bf{D}}$  & 61.55 & 80.69 & 55.96  & 73.23 \\
        w/ shared ${\bf{D}}$ (Ours) &{\bfseries 64.28} &{\bfseries 81.95} &{\bfseries 57.52} &{\bfseries 74.84}\\
    \hline
\end{tabular}
\label{tab:s_dic}
\end{table}

\begin{table}
\centering
\caption{Comparisons between different types of prototypes ${\bf{P}}$ in PGU, including the modality shared prototypes (w/ shared ${\bf{P}}$) and the modality unshared prototypes (w/ unshared ${\bf{P}}$).}
\begin{tabular}{c|cc|cc}
    \hline
      \multirow{2}*{Type of ${\bf{P}}$} & \multicolumn{2}{c|}{CUHK-PEDES} & \multicolumn{2}{c}{ICFG-PEDES}\\
    \cline{2-5}
       & \multicolumn{1}{c}{Rank-1} & \multicolumn{1}{c|}{Rank-5}& \multicolumn{1}{c}{Rank-1} & \multicolumn{1}{c}{Rank-5}\\
    \hline
        w/ unshared ${\bf{P}}$ & 63.78  & 82.15 & 57.73  & 75.11\\
        w/ shared ${\bf{P}}$ (Ours)  &{\bfseries 65.25} &{\bfseries 83.10} &{\bfseries 59.02} &{\bfseries 75.32} \\
    \hline
\end{tabular}
\label{tab:s_pro}
\end{table}

\subsection{Comparisons on Time Complexity}
As discussed in Section~\ref{sec:setion3.3}, our LGUR has advantages in time complexity.
In this subsection, we evaluate the training time\footnote{The training time refers to the average time taken to process one mini-batch of images and descriptions.}, inference time\footnote{The inference time includes the feature extraction time for a given query and time for similarity computation with all gallery images.} and Rank-1 accuracy of LGUR, three works that do not implement the cross-modal attention mechanism (\emph{i.e.}, Dual Path \cite{zheng2020dual}, CMPM/C \cite{zhang2018deep}, and SSAN \cite{ding2021semantically}), and another three methods that adopt cross-attention mechanism (\emph{i.e.}, MIA \cite{niu2020improving}, SCAN \cite{lee2018stacked}, and NAFS \cite{gao2021contextual}).
To facilitate fair comparison, we set the image size of input images to 384 $\times$ 128 pixels and the batch size to 64 for all methods.
All experiments are conducted on a Titan X GPU.
As shown in Table~\ref{tab:costtime}, LGUR is dramatically more efficient than all methods that rely on cross-modal attention operations.
This advantage in efficiency benefits from the disentanglement of the textual and visual feature extraction in LGUR.
In addition, the computational cost of LGUR is competitive with the works that do not incorporate cross-modal attention mechanisms.
For example, LGUR costs 26ms per query on the CUHK-PEDES database, which is faster than SSAN at 76ms.
Considering the trade-off between accuracy and efficiency, LGUR is thus superior to other methods.

\subsection{Ablation Study}
In this subsection, we analyze the effectiveness of each key component in the LGUR framework. Here, we adopt the DeiT-Small as the visual backbone.

\noindent\textbf{Effectiveness of PGU.}
In Table~\ref{tab:ablation}, the efficacy of PGU is revealed via the experimental results of No.0 \emph{v.s.} No.1. Adding PGU on the baseline promotes the Rank-1 accuracy of the baseline by 4.59\% on CUHK-PEDES. When comparing the results of No.2 and No.6 experiments, PGU can improve the Rank-1 accuracy from 61.86\% to 65.25\% on baseline+DGA. The above results clearly show that the unified textual and visual feature representations from PGU are beneficial for improving the performance.

\noindent\textbf{Effectiveness of DGA.}
The experimental results of No.0 \emph{v.s.} No.2, and No.1 \emph{v.s.} No.6 in Table~\ref{tab:ablation} demonstrate the efficacy of DGA.
In particular, when adding DGA to the baseline, the Rank-1 accuracy is promoted by 3.19\% and 3.74\% on CUHK-PEDES and ICFG-PEDES, respectively. 
These results justify that DGA well aligns the feature granularity of the two modalities and therefore promotes the retrieval accuracy.

Meanwhile, we provide comprehensive experimental analysis to further explore the impact of each component in DGA.

First, the most important part in DGA is the multi-modality shared dictionary (${\bf{D}}$). In Table~\ref{tab:ablation}, when adopting ${\bf{D}}$ on baseline+PGU in experiment No.3, the performance is promoted by 1.02\% on CUHK-PEDES, showing that the reconstructed features from ${\bf{D}}$ is more effective. This is because via the shared ${\bf{D}}$ for the visual and textural features, their granularity is roughly aligned.

\begin{figure}[t]
  \centering
  \includegraphics[width=\linewidth]{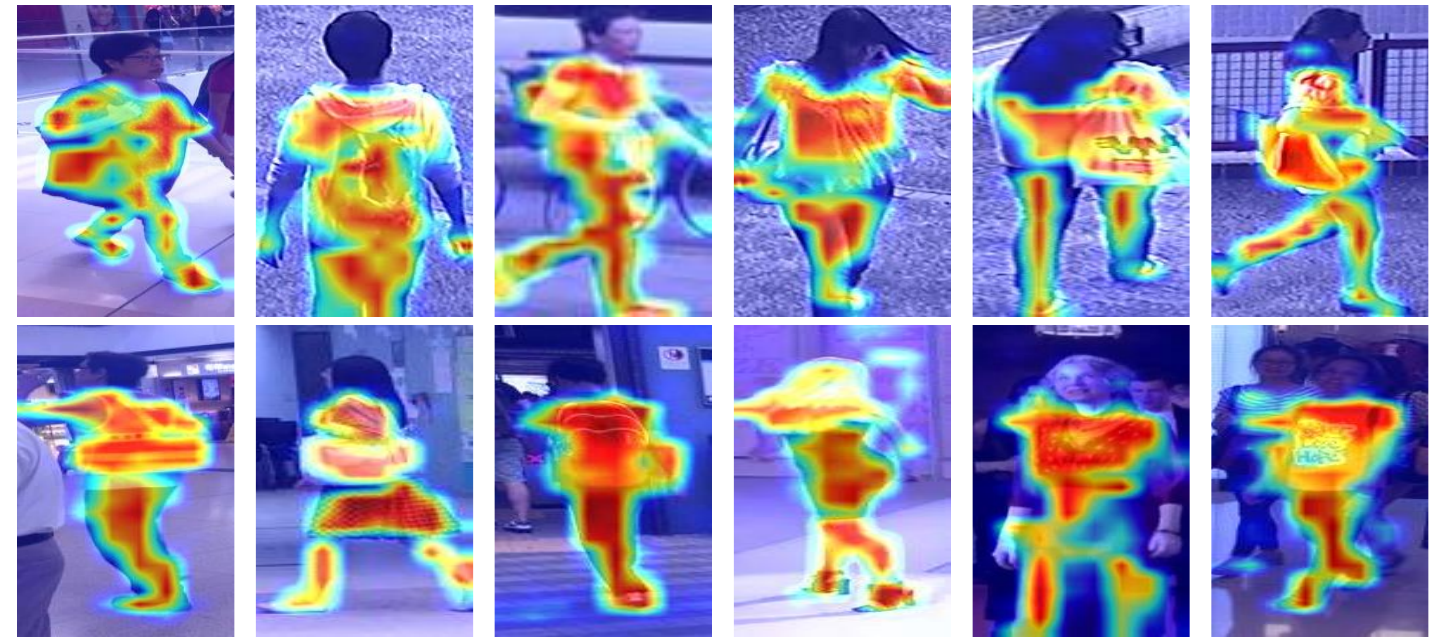}
  \caption{Visualization of the foreground masks ${\bf{M}}$. The pedestrian body areas are highlighted while the cluttered background is suppressed.}
  \label{fig:mask}
\end{figure}

Second, we also enable DGA to focus on the reconstruction of foreground visual features. As shown in experiments No.3 and No.4 in Table~\ref{tab:ablation}, the foreground mask $\bf{M}$ further promotes the Rank-1 accuracy from 64.28\% to 64.64\% on CUHK-PEDES. This result reveals that foreground-oriented reconstruction helps to further reduce the modality gap and relieves the optimization burden of ${\bf{D}}$, as analyzed in Section~\ref{sec:setion3.2}. Meanwhile, we also visualize the foreground mask $\bf{M}$ in Figure~\ref{fig:mask}. It clearly shows that $\bf{M}$ makes the model to focus on the meaningful pedestrian body instead of the useless background.

Third, we adopt ${\bf{V}}_{g}$ in Eq.~\eqref{eq:visual_foreground_guidance} without $\bf{M}$ to guide ${\bf{D}}$ to learn atoms of coarse granularities as the same as that in text. As shown by the experimental results No.5 \emph{v.s.} No.3, the Rank-1 accuracy improves by 0.65\% on ICFG-PEDES when ${\bf{V}}_{g}$ is adopted, verifying the effectiveness of the guidance from ${\bf{V}}_{g}$.

Finally, our LGUR combines ${\bf{D}}$, $\bf{M}$, $\emph{G}$ and PGU together, achieving the best result.
For example, LGUR outperforms the baseline by as much as 6.58\% and 6.93\% in Rank-1 accuracy on CUHK-PEDES and ICFG-PEDES, respectively.

\noindent\textbf{Comparisons with variants of MSD.}
We adopt the multi-modality shared dictionary (${\bf{D}}$) to reconstruct both the textual and the visual features.
The experiment No.3 in Table~\ref{tab:ablation} has justified the effectiveness of ${\bf{D}}$.
In Table~\ref{tab:s_dic}, we further verify its design via comparison with several possible variants.

First, MSD significantly outperforms the naïve self-attentions layer (SA) in transformer~\cite{vaswani2017attention}. 
The result in Table~\ref{tab:s_dic} reveals that reconstruction via ${\bf{D}}$ is more effective since it aligns the granulary of both modalities. In comparison, the self-attention layer cannot explicitly reduce the granularity gap between the two modalities, and thus is inferior to our MSD.
Second, the shared ${\bf{D}}$ between the two modalities significantly outperforms its unshared counterpart,
as shown in the last two rows in Table~\ref{tab:s_dic}. Actually, the separate ${\bf{D}}$ for the two modalities may enlarge the modality gap, which hurts the ReID performance.
Overall, the design of our shared ${\bf{D}}$ in MSD is a good choice to narrow down the modality gap.

\noindent\textbf{Comparisons with variants of PGU.}
The prototypes ${\bf{P}}$ in PGU are shared between the image and text modalities.
To validate the advantage of this design, we also evaluate the unshared ${\bf{P}}$, 
\emph{i.e.}, using different ${\bf{P}}$ for text and image.
In Table~\ref{tab:s_pro}, it is clear that adopting shared ${\bf{P}}$ outperforms the unshared ${\bf{P}}$ by large margins.
It is due to the superiority of the shared ${\bf{P}}$, which can improve the granularity unification between the two modalities.

\subsection{Qualitative Results}
\begin{figure}[t!]
\begin{center}
\includegraphics[width=\linewidth]{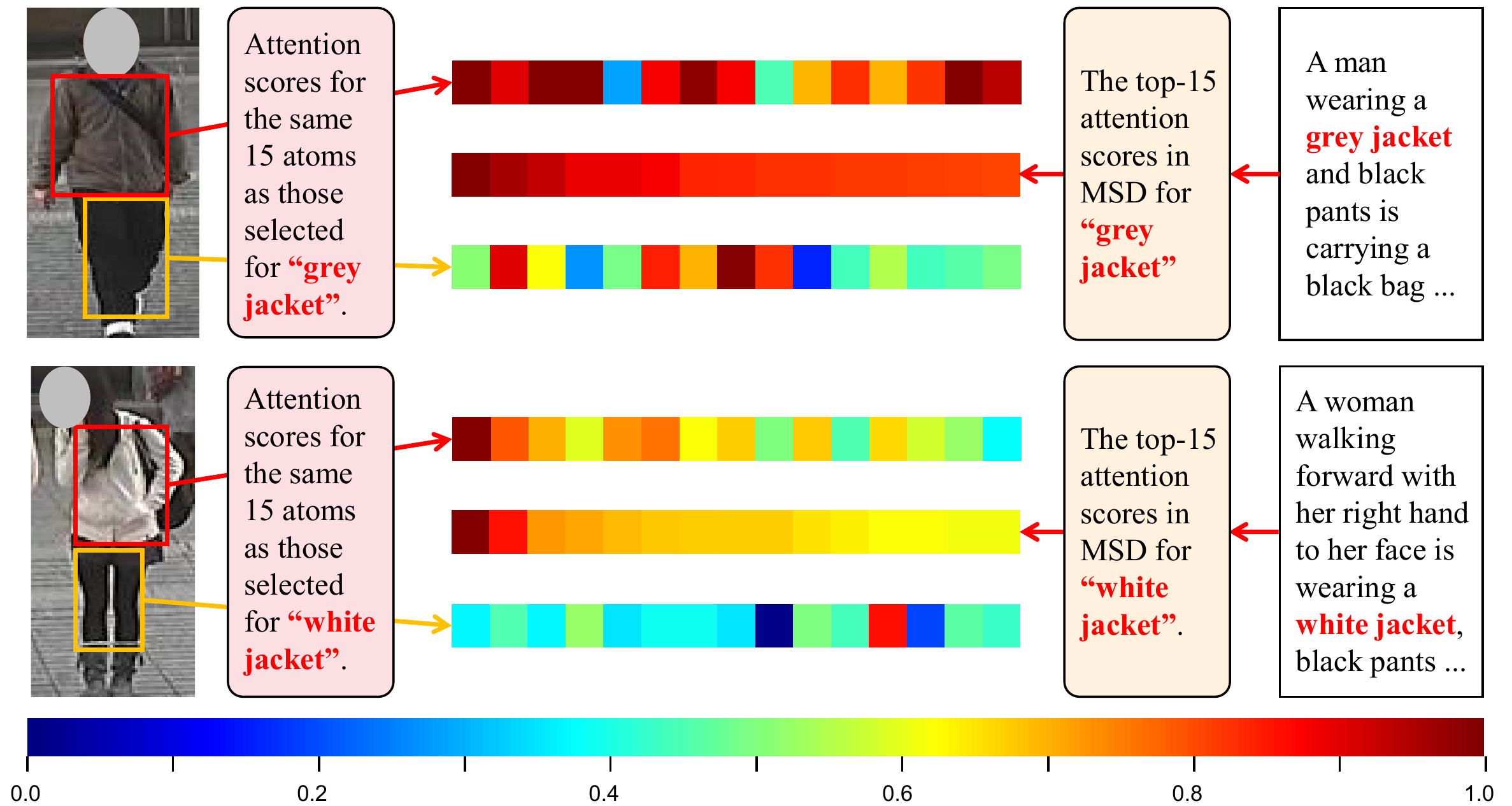}
\end{center}
  \caption{Visualization of the attentions scores between one selected phrase with ${\bf{D}}$ (the right part) and the image patch with ${\bf{D}}$ (the left part) in the DGA module. The red rectangle denotes the text-matched image patch, while the yellow one represents the text-irrelevant image patch.
  }
\label{fig:view_1}
\end{figure}

Here, we provide qualitative results to demonstrate MSD’s reconstruction ability in Figure~\ref{fig:view_1}. Ideally, if one phrase describes an image patch, their cross-attention maps with ${\bf{D}}$ should be similar.
Otherwise, their cross-attention maps should be different.
Considering that the whole attention map is large, here we only show 15 selected attention scores.

Specifically, we first choose one phrase and its corresponding image patch, which are highlighted in red text colour and red rectangle in Figure~\ref{fig:view_1}, respectively. 
We also select one irrelevant image patch to the phrase and frame the patch in yellow. 
We first draw the top-15 attention scores that represent the highest similarities between the phrase and ${\bf{D}}$.
Then we show the attention scores on the same 15 atoms as above between each of the two patches and ${\bf{D}}$.
Results in Figure~\ref{fig:view_1} show that the scores for the matched phrase-patch pair are similar, which indicates that the atoms in ${\bf{D}}$ can well represent features for both text and image.
In contrast, the attention scores for the irrelevant phrase-patch pair are different. This phenomenon indicates that the granularity gap of the two modalities can be reduced after the dictionary-based reconstruction.

\section{Conclusion}
In this paper, we have introduced a novel framework named LGUR to learn granularity-unified representations for the text-to-image ReID task. This framework includes a Dictionary-based Granularity Alignment (DGA) module and a Prototype-based Granularity Unification (PGU) module. In the DGA module, we build a Multi-modality Shared Dictionary (MSD) to reconstruct both visual and textual features, such that their granularity can be unified. We further provide a cross-modal guidance strategy and a foreground mask to facilitate the optimization of MSD parameters. In the PGU module, we adopt a set of shared prototypes for diverse textual and visual feature extraction, which further aligns the granularity of both modalities. 
Extensive experiments on two large-scale databases demonstrate the effectiveness of our LGUR.

\begin{acks}

This work was supported by the CCF-Baidu Open Fund, the National Natural Science Foundation of China under Grant 62076101 and 61702193, the Program for Guangdong Introducing Innovative and Entrepreneurial Teams under Grant 2017ZT07X183, Guangdong Basic and Applied Basic Research Foundation under Grant 2022A1515011549, and Guangdong Provincial Key Laboratory of Human Digital Twin (2022B1212010004).
\end{acks}

\bibliographystyle{ACM-Reference-Format}
\bibliography{sample-base}

\appendix

\end{document}